\documentclass[conference]{IEEEtran}
\IEEEoverridecommandlockouts
\usepackage{cite}
\usepackage{url}
\usepackage{amsmath,amssymb,amsfonts}
\usepackage{amsthm}
\usepackage{graphicx}
\usepackage{textcomp}
\usepackage{xcolor}
\usepackage{multirow}
\usepackage[linesnumbered,ruled,vlined]{algorithm2e}
\usepackage{algpseudocode}
\let\oldnl\nl
\newcommand{\nonl}{\renewcommand{\nl}{\let\nl\oldnl}}

\newtheorem{definition}{Definition}
\begin{document}

\title{Out-of-Distribution Detection for Neurosymbolic Autonomous Cyber Agents \\
}

\author{\IEEEauthorblockN{Ankita Samaddar, Nicholas Potteiger, Xenofon Koutsoukos}
\IEEEauthorblockA{\textit{Department of Computer Science} \\
\textit{Vanderbilt University}\\
Nashville, TN, USA \\
\{ankita.samaddar, nicholas.potteiger, xenofon.koutsoukos\}@vanderbilt.edu}}

\maketitle

\begin{abstract}
Autonomous agents for cyber applications take advantage of modern defense techniques by adopting intelligent agents with conventional and learning-enabled components. These intelligent agents are trained via reinforcement learning (RL) algorithms, and can learn, adapt to, reason about and deploy security rules to defend networked computer systems while maintaining critical operational workflows. However, the knowledge available during training about the state of the operational network and its environment may be limited.
The agents should be trustworthy so that they can reliably detect situations they cannot handle, and hand them over to cyber experts.  In this work, we develop an out-of-distribution (OOD) Monitoring algorithm that uses a Probabilistic Neural Network (PNN) to detect anomalous or OOD situations of RL-based agents with discrete states and discrete actions. To demonstrate the effectiveness of the proposed approach, we integrate the OOD monitoring algorithm with a neurosymbolic autonomous cyber agent that uses behavior trees with learning-enabled components. We evaluate the proposed approach in a simulated cyber environment under different adversarial strategies. Experimental results over a large number of episodes illustrate the overall efficiency of our proposed approach.

\end{abstract}


\section{Introduction}\label{sec:intro}
\noindent
Autonomous agents for cyber applications require to learn and deploy security rules to defend cyber-attacks without any human intervention. Defending a network from cyber-attacks needs constant monitoring of the system and selecting the appropriate actions whenever a security breach is detected while still maintaining the critical operational workflows. Existing security standards often fall short in designing these cyber-defense agents against sophisticated adversarial attacks. Hence, a combination of security standards along with learning enabled components (LECs) that can learn, detect and defend the system from attacks are needed. These LECs are typically function approximators with a reinforcement learning (RL) policy,
so that they can take optimal actions and effectively mitigate dynamic complex attacks. 
However, uncertainties pose a significant challenge in characterizing the trustworthiness of these autonomous agents. These uncertainties may arise due to limited knowledge available to the autonomous agents about the runtime behavior of the operational system and environment at the time of designing or training these agents. The consequences can propagate deep into the system and can impact system behaviors at all levels. Thus, anomaly or out-of-distribution (OOD) detection methods need to be incorporated to identify information that is nonconformal with the environment used in training. 

In a cyber environment, a trustworthy autonomous agent should not only generate optimal actions in known situations that it understands, but also reliably detect situations that are nonconformal, reject them, and pass them over to cyber experts. For instance, an autonomous cyber-defense agent trained against different types of denial-of-service (DoS) attacks such as blocking the traffic or blocking the IP address of a specific host, can detect and handle such situations promptly. However, if a new type of attack such as an unauthorized access to the network server due to authentication breaches, emerges into the system, then the autonomous cyber-defense agent should detect such situations as anomalous or nonconformal so that they can be handled by cyber experts. Therefore, a safety assurance method needs to be associated with an autonomous agent to make it trustworthy under every situation. Anomaly or OOD detection is an established area of research in robotic systems~\cite{farid2021taskdriven}, cyber-physical systems~\cite{shreyas2022, Cai2020iccps}, etc. However, OOD detection to defend cyber-attacks in autonomous networks is not well explored in the literature. With this objective into consideration, we aim to design an OOD Monitoring algorithm to make autonomous cyber-defense agents trustworthy. Apart from monitoring cyber-defense agents for autonomous networks, our proposed algorithm can also be integrated with any RL-based agent with discrete states and discrete actions.

Although different RL policies are used in designing cyber-defense agents to defend autonomous networks, they fail to scale and optimize these systems as the networks grow and become more complex over time. 
To overcome these challenges, in our prior work, we 
proposed a neurosymbolic model representation of an autonomous cyber-defense agent using behavior trees (BTs)~\cite{Colledanchise_2018} with LECs, more specifically known as the Evolving Behavior Trees (EBTs)~\cite{potteiger2024design}. The EBT structures are modular in nature with capabilities to adapt to multiple dynamic attack situations. They can capture an explicit hierarchy of subtasks and control flows, and can deploy LECs to execute specific subtasks. Their generalizable structures make them deployable to a real system as well as in simulation. However, an online monitoring technique needs to be incorporated with the EBT-based autonomous cyber-defense agents to ensure safety by detecting OOD situations of the system at runtime. As a solution to this problem, we integrate our proposed OOD Monitoring algorithm with our EBT-based autonomous cyber-defense agent and demonstrate its effectiveness.

Thus, the main contributions of our work are as follows.
\begin{enumerate}
    \item We propose an OOD Monitoring algorithm for RL-based agents with discrete states and discrete actions. The algorithm uses a Probabilistic Neural Network (PNN)~\cite{hajdarevic2015pnn} to capture the in distribution behavior of the system and detects OOD situations when the system deviates from the expected behavior.
    \item We integrate the OOD Monitoring algorithm with an EBT-based autonomous cyber-defense agent to demonstrate its ability to detect OOD situations in autonomous networks at runtime and take mitigation actions that improve the overall performance.
    \item We perform a comprehensive evaluation using CybORG CAGE Challenge Scenario 2, a complex network simulation environment~\cite{cage_challenge_2}. We evaluate the proposed approach over multiple episodes under different adversarial strategies. Experimental results under different settings illustrate the efficacy of our proposed solution.
\end{enumerate}

\section{Related Work}\label{sec:related}
\noindent
Traditional security measures are not sufficient to defend autonomous networks from sophisticated cyber-attacks. As a result, nowadays, different RL methods are deployed to develop more advanced and interpretable learning enabled defense policies in autonomous networks. CybORG serves as a popular cyber security research environment for training and development of different RL-based autonomous agents~\cite{Kiely2023OnAA}. Some RL-based agents utilize a goal-conditioned hierarchical RL (HRL) to validate trained defense strategies in CybORG~\cite{autonomous2022foley, Wolk2022BeyondCI}, whereas, others use an ensemble approach aggregating policy outputs~\cite{MolinaMarkham2021NetworkED}. For emulating cyber-attacks and defense scenarios, Markham \emph{et al.} developed a novel tool called FARLAND that focuses on  realistic cyber-defense environments and curriculum learning for cyber-defense agents~\cite{MolinaMarkham2021NetworkED}. 

An emerging area of research in designing autonomous cyber-defense is to use Neurosymbolic AI that combines pattern recognition capabilities of neural networks along with explicit reasoning of symbolic systems~\cite{neuro2023milcom}. An effective way to design these neurosymbolic autonomous agents is to use behavior trees (BTs)~\cite{Colledanchise_2018}. RL or HRL techniques are used to learn, jointly optimize and generate policies to capture complex BT behaviors~\cite{Lundberg1640875, Li2021MixedDR, Bacon2017AAAI}. Our prior work proposed an approach that uses genetic programming to construct EBTs with LECs that analyzes the system behavior and apply appropriate mitigation strategies against adversarial attacks to ensure autonomous cyber-defense in enterprise networks~\cite{potteiger2024design}. However, none of these works can ensure safety of the system at runtime by detecting OOD behavior of the autonomous cyber-defense agents.  

OOD detection is well studied in the literature for 
safety critical applications such as autonomous vehicles~\cite{Filos2020CanAV}, robotics~\cite{farid2021taskdriven}, etc. Cai \emph{et al.} proposed an OOD detection approach where they used variational autoencoders and deep support vector data description to learn the system and use them in real-time to compute the nonconformity of new inputs relative to the training set in advanced emergency braking system and a self-driving end-to-end controller~\cite{Cai2020iccps}. Ramakrishna \emph{et al.} designed a $\beta$-variational autoencoder detector with partially disentangled latent space to detect OOD scenarios with variations in the image features~\cite{shreyas2022}. Farid \emph{et al.} presented a Probably Approximately Correct (PAC)
Bayes framework to train policies for a robotic environment with guaranteed bounds on performance on the training data distribution and detects OOD behavior of the robot by capturing the violation of the performance bound on the test environment~\cite{farid2021taskdriven}. Averly \emph{et al.} presented a unifying framework to detect OOD scenarios caused by both semantic and covariate shifts in uncontrolled environments for a variety of models~\cite{Averly2023UnifiedOD}. Yang \emph{et al.} presented a full-spectrum OOD detection model that uses a simple feature-based semantics score function to account for semantic shifts and become tolerant to covariate shifts in image data~\cite{yangfull2023}. 
However, none of these works focus on
OOD detection scenarios for RL agent based autonomous systems.

A few works on different OOD detection approaches for RL-based agents exist in the literature~\cite{amaas2023rl, ood2024aamas, singh2024amaas}. 
However, all of these works are applicable for continuous state space. The autonomous cyber-defense agent considered in our work consists of discrete and partially observable states and discrete actions. Hence, existing approaches for OOD detection for RL agents are not directly applicable to our system. In this work, we develop an OOD Monitoring algorithm that can detect OOD scenarios in autonomous networks to assure safety of our system at runtime.

\begin{figure*}[t]
    \centering
    \includegraphics[width=0.7\linewidth]{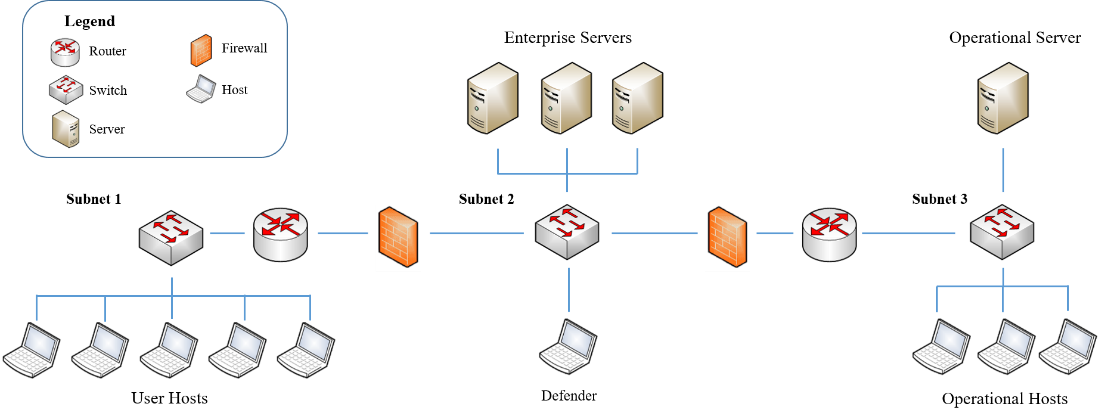}
    \caption{Network architecture of CybORG Cage Challenge Scenario 2 with three subnets~\cite{cage_challenge_2}; Subnet~1 with five hosts, Subnet~2 with three enterprise servers and a defender host, Subnet~3 with an operational server and three operational hosts.}
    \label{fig:cagechallenge2}
\end{figure*}
\section{Autonomous Agents for Cyber-Defense}\label{sec:back}
\noindent
In our prior work, we developed a robust autonomous cyber-defense agent that interacts with the environment and uses cyber-agent actions against dynamic cyber-attacks~\cite{potteiger2024design}. 
We evaluated our agent in CybORG, a complex network simulation environment that abstracts real world scenarios~\cite{cyborg_acd_2021}. We considered the network scenario presented in CAGE Challenge Scenario 2~\cite{cage_challenge_2}. Fig.~\ref{fig:cagechallenge2} shows the network architecture. The network comprises three subnets: 
\begin{enumerate}
    \item Subnet~1 with five non-critical user hosts.
    \item Subnet~2 with three enterprise servers that support the activities of the hosts in Subnet~1 and a host that acts as the defender.
    \item Subnet~3 with three operational hosts and a critical operational server that is responsible to ensure that the network is functioning properly.
\end{enumerate}

CybORG interface can be used to construct and evaluate the attacker (red agent) and the defender (blue agent) using LECs. Each scenario run in CybORG consists of a fixed number of timesteps over a fixed period of time. In every timestep, the red and the blue agents each chooses and executes an action from a set of available actions. The red agent starts each scenario run with an initial access to one of the user machines in Subnet~1. Thereafter, the red agent can scan hosts and subnets in the enterprise network, exploit the hosts and perform privilege escalation. Once the red agent has exploited the enterprise server and accessed the IP address of the operational server, it gains access to the operational network. The operational server provides a service to maintain the system owner's operations. The main target of the red agent is to disrupt this service through the ``Impact" action as long as possible. 

To mitigate red agent actions in each timestep, the blue agent may take no action (Sleep), monitor the network connections and malicious processes (Monitor), analyze information on files that are associated with a host or a server (Analyze), deploy one of the seven decoys on a host or a server if a red agent accesses a new service (Deploy Decoy), remove any malicious files, processes or services from a host or a server (Remove), and restores a host or a server to a known safe state (Restore). 

Cage Challenge Scenario 2 presents two types of red agent strategies. The $Meander$ agent, that exploits each subnet one by one by seeking and gaining privileged access on all hosts in a subnet before moving on to the next subnet, finally reaching the operational server. The $B\_line$ agent, that uses full knowledge of the network and directly traverses to the critical operational server. In our prior work, a third red agent strategy, the $RedSwitch$ was introduced that combines $B\_line$ and $Meander$~\cite{potteiger2024design}. The $RedSwitch$ strategy first instantiates a red agent using $Meander$ strategy, and after a random number of timesteps it switches the red agent strategy to $B\_line$ strategy.

\begin{figure*}[h]
    \centering
    \includegraphics[width=\linewidth]{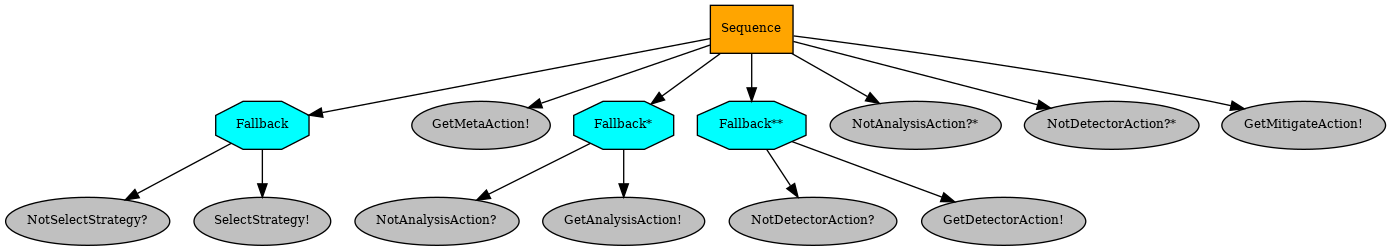}
    \caption{Optimal Behavior Tree for Autonomous Cyber-Defense~\cite{potteiger2024design}}
    \label{fig:gpbt}
\end{figure*}
\section{Evolving Behavior Tree based Autonomous Cyber-Defense Agent.}\label{sec:agent}
\noindent
Neurosymbolic AI can be leveraged in cyber-security to learn the system behavior holistically to mitigate sophisticated adversarial attacks. 
Given a goal or specification, a symbolic structure is used as a model that interacts with the environment and selects appropriate actions against adversarial red agents. We use behavior trees (BTs) as the symbolic structure in the design of our agent because of their capabilities to integrate LECs that allow us to learn and reason about cyber-defense control at a high level and adapt to environmental shifts. Their modular structures allow us to integrate new capabilities into the system. Moreover, they are generalizable, that allow us to map them to both abstract and realistic environments.



In our prior work, we presented a neurosymbolic approach using Evolving Behavior Trees (EBTs) to develop a robust autonomous cyber-defense agent that interacts with the environment and uses cyber-agent actions against dynamic cyber-attacks~\cite{potteiger2024design}. We abstracted our EBT-based cyber-defense agent from a pursuit evasion game environment using genetic programming and evaluated our agent in CybORG CAGE Challenge Scenario 2~\cite{potteiger2024design}. Fig.~\ref{fig:gpbt} shows the optimal BT structure for our autonomous cyber-defense agent.

In the execution of a BT, each timestep is called a tick. A BT starts executing from the root node and follows a Depth-First Traversal. On completing execution of a behavior in a BT node, the child returns a status of $Running$ if its execution is underway, $Success$ if it has achieved its goal, and $Failure$ otherwise. The behaviors in a BT can be classified into two groups - $Control$ behaviors and $Execution$ behaviors. $Control$ behaviors are the internal behaviors in a BT that control the logical flow of switching between the behaviors. $Execution$ behaviors are the leaf behaviors in a BT that execute specific action in the environment. $Control$ behaviors in a BT can be either $Sequence$ or $Fallback$. $Sequence$ executes a set of child behaviors sequentially until all children return $Success$, otherwise returns $Failure$. $Fallback$ executes the leaf behaviors until one child returns $Success$, otherwise returns $Failure$. $Execution$ behaviors in a BT can be a $Condition$ or an $Action$ behavior, the return status of which is dependent on the intended logical condition or user-defined functionality. Each tick of the BT comprises execution of the entire BT from the root in a depth-first manner till the status from the leaf behaviors are returned, after which it propagates back upto the root recursively by updating the status of the parent control behaviors.

From Fig.~\ref{fig:gpbt}, we define five $Action$ behaviors in our autonomous cyber-defense agent. 
\begin{enumerate}
    \item $SelectStrategy!$ selects a defense strategy depending on the adversarial (red agent) movement.
    \item $GetMetaAction!$ selects one of the three defense behaviors based on the defense (blue agent) strategy.
    \item $GetAnalysisAction!$ monitors or  analyzes the environment by retrieving new information that is unknown to the
    blue agent.
    \item $GetDetectorAction!$ deploys a detection mechanism in the environment to alert an adversarial activity.
    \item $GetMitigateAction!$ prevents an adversary from achieving their objective, e.g., blocking an adversarial movement or restoring the network host/server to a previously ``safe" state.
\end{enumerate}

Besides the $Action$ behaviors, there are four $Condition$ behaviors. One $Condition$ behavior for $SelectStrategy!$ is to ensure that the correct defense strategy is selected or shifted in every tick. Remaining three $Condition$ behaviors for each of the three defense behaviors is to ensure that the correct behavior is enacted when chosen by $GetMetaAction!$.

These five cyber-defense actions are mapped to their appropriate behaviors in CybORG CAGE Challenge Scenario 2. $GetAnalysisAction!$ behavior maps to $Analyze$ and $Monitor$ actions. $GetDetectorAction!$ maps to a greedy deterministic policy that selects a $Deploy Decoy$ action on a host or a server. $GetMitigateAction!$ maps to $Remove$ and
$Restore$ actions to prevent the adversary from causing further damage. To select one of the above $Action$ behaviors, a cyber-agent controller policy is executed by the $GetMetaAction!$ behavior. Finally, the type of controller policy to be used is determined by the $SelectStrategy!$ behavior depending on the red agent movements, and the $NotSelectStrategy?$ behavior determines if the controller policy needs to be switched.

Our EBT-based cyber-defense agents are capable of defending a variety of cyber-attacks, however, they need to be monitored at every timestep so that any anomaly or nonconformity in their behavior is detected and handled by cyber experts. This motivates us to propose an out-of-distribution (OOD) Monitoring algorithm to detect any deviation in the expected behavior of our autonomous cyber-defense agents.

\section{Out-of-distribution Detection}\label{sec:runtimemonitor}
\noindent
Although our EBT-based autonomous cyber-defense agents are robust against dynamic cyber-attacks, uncertainties introduce a significant challenge for characterizing trustworthiness of these agents. These uncertainties arise from knowledge gaps about the runtime state of the system and the environment at the time of design and training these agents. As a consequence, the system behavior may get impacted at all levels that may lead the system to unsafe states. Thus, the main objective of this work is to design a trustworthy autonomous agent that can reliably detect anomaly or out-of-distribution situations and hand them over to cyber-experts to assure safety of the system at runtime. 

\subsection{Problem Formulation}\label{subsec:problem}
\noindent
Our system can be modeled as a decision-making system with RL agents that sequentially interacts with the environment by executing actions. At each timestep, our system transitions from one state to another based on the red agent (adversarial) and blue agent (defender) actions. However, the red agent actions are not observable. Thus, our system can be formally represented by a discrete-time Partially Observable Markov Decision Process (POMDP), $\mathbb{M} := (S, A, T, R, \mu_0)$~\cite{mdp}. $S$ denotes the set of states or observations which are discrete and partially observable, $A$ denotes the set of defender (blue agent) actions which are discrete, $T$ denotes the conditional transition probabilities, $R : S \times A \times S \mapsto R$ denotes the reward function, and $\mu_0 : (s_0, a_0)$, $s_0 \in S$, $a_0 \in A$, denotes the initial state and action. At each timestep $t-1$, the defender (blue agent) takes an action $a_{t-1} \in A$ which causes the system to transition from $s_{t-1}$ to $s_{t}$ with probability $T(s_{t}|s_{t-1},a_{t-1})$ and gets a reward $r_{t-1}$. The objective of the blue agent is to select actions at each timestep so that the cumulative rewards maximize over time, \emph{i.e.}, $\sum_{t=1}^{t \rightarrow \infty} r_{t-1}$.

We define \emph{Transition Probability Threshold} to quantify out-of-distribution situations of our system.

\begin{definition}
    Given a neurosymbolic cyber-agent trained with a policy $\pi$, a state transition at timestep $t-1$, denoted by $(s_{t-1},a_{t-1}) \rightarrow s_{t}$, is considered to be an out-of-distribution (OOD) transition based on policy $\pi$ if the probability of occurrence of this transition in the training data is less than threshold $\rho$, \emph{i.e.}, $\Pr\left((s_{t-1},a_{t-1}) \rightarrow s_{t} \right) < \rho$. We refer $\rho$ as the Transition Probability Threshold.
\end{definition}

\vspace{0.3em}
\noindent
\emph{Problem Statement : Given a network consisting of hosts, enterprise servers and operational servers (as shown in Fig.~\ref{fig:cagechallenge2}) and a neurosymbolic cyber-agent trained with a policy $\pi$, our objective is to develop a safety assurance algorithm to detect shifts from the distribution used for training.}

In this work, we specifically address two key questions.

\begin{enumerate}
    \item \emph{Can we assure safety if the system transitions to any state $s'$ such that $\Pr\left((s,a) \rightarrow s'\right) < \rho$ in our training distribution?}
    \item \emph{Can we assure safety if the red agent switches to a different strategy than the one used for training?}
\end{enumerate}

\subsection{OOD Monitoring Algorithm}\label{subsec:runtime}
\noindent
We develop an OOD Monitoring algorithm that executes at every timestep to detect any deviations of the current observation from the one used for training the autonomous agent. Haider \emph{et al.} presented a model-based OOD detection framework for RL agents using probabilistic dynamics model~\cite{ood2024aamas}. Unlike \cite{ood2024aamas}, the states in our system are discrete and partially observable. Hence, we cannot directly apply their framework to our system. We propose to use a Probabilistic Neural Network (PNN) to learn the dynamics of our system that is characterized by non-deterministic transitions of the partially observable states~\cite{hajdarevic2015pnn}.


Our OOD Monitoring algorithm consists of three phases.

\begin{itemize}
    \item An \emph{Data Generation} phase
    \item An \emph{Training} phase
    \item An \emph{OOD Monitoring} phase 
\end{itemize}

Algorithm~\ref{alg:monitor} shows the main steps of our proposed approach.

\vspace{0.3em}
\noindent
\textbf{Data Generation Phase:} 
To learn the dynamics of our system, we need to learn the dynamic function that characterize the transition probabilities of our system. In order to achieve this, we need to collect data, \emph{i.e.}, states, actions and their transitions $(s_{t-1},a_{t-1}) \rightarrow s_{t}$, by interacting with our system, and then generate a PNN with the collected data. 

Given a trained blue agent control policy $\pi$ for a fixed red agent strategy, we collect transitions $(s_{t-1}, a_{t-1}) \rightarrow s_{t}$  for $\tau$ timesteps, ($\tau$ is very large), over multiple episodes (say $N$) and generate the training data $D_{train}$.

\vspace{0.3em}
\noindent
\textbf{Training Phase:} Let us assume that $f_{\theta}$ represent the discrete dynamic function that characterize our decision making system. Since our system is non-deterministic, it exhibits different behaviors on different runs even with the same inputs. Thus, $f_{\theta}$ can be formally represented as a mapping of the previous state and action to a set of possible $k$ current states, where $k$ is any positive number, \emph{i.e.}, $f_{\theta}(s_{t-1},a_{t-1}) = \{s^1_{t}, s^2_{t},\ldots,s^k_{t}\}$. Each of these $s^i_{t}$ is associated with some probability conditioned on $(s_{t-1},a_{t-1})$ and generated from the observed samples in the training data.

\begin{figure}
    \centering
    \includegraphics[width=\linewidth]{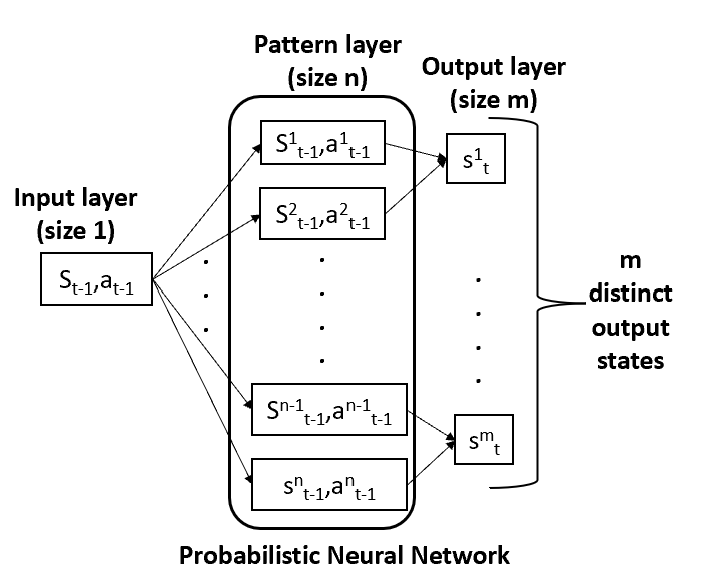}
    \caption{A PNN with input layer of size 1, pattern layer of size $n$ = $N \times \tau$, and output layer of size $m$ where $m$ denotes distinct observed states in $D_{train}$.}
    \label{fig:pnn}
\end{figure}

To learn $f_{\theta}$, we develop a PNN where the input layer is of size~1, the pattern layer is of size equal to the size of the training data, \emph{i.e.}, $N \times \tau$, and the output layer is of size $m$ (say) where $m$ is equal to the number of distinct observed states in the training data. Unlike the generic structure of a PNN which consists of four layers~\cite{hajdarevic2015pnn}, 
our PNN model does not have the fourth layer, \emph{i.e.}, the decision layer. Instead, the summation layer serves as our output layer. Fig.~\ref{fig:pnn} shows a schematic diagram of our PNN model. For each input, $(s_{t-1}, a_{t-1})$, at the input layer, the PNN matches the input with all the entries in the pattern layer.
The PNN activates only those entries ($s^i_{t}$) in the output layer whose previous states and actions match with the input data.

\vspace{0.3em}
\noindent
\textbf{OOD Monitoring Phase: } Given our decision making system trained with an agent policy $\pi$ and a trained PNN for the same policy, let the system transition from $s_{t-1}$ to $s_t$ due to execution of action $a_{t-1}$ at timestep $t-1$ following control policy $\pi$. To detect OOD behavior of our system, we feed the same previous state and action ($s_{t-1}$, $a_{t-1}$) to the PNN that generates a set of $k$ predicted current states, $\{s^1_t, s^2_t,\dots,s^k_t\}$ (say). For each $s^i_t$ in the set of predicted current states, the associated transition probability $\Pr\left((s_{t-1},a_{t-1}) \rightarrow s^i_t \right)$ is calculated from the activated connections in the PNN. We consider the current state $s_t$ of the system to be in-distribution (ID), if $s_t \in \{s^1_t, s^2_t,\dots,s^k_t\}$ and the associated transition probability $\Pr\left((s_{t-1},a_{t-1}) \rightarrow s_t \right)$ is greater than Transition Probability Threshold, $\rho$. Otherwise, $s_t$ is considered to be out-of-distribution (OOD). This is because, if a particular state transition occurs in our training distribution for a significantly large number of times, then we acquire high confidence about the state and its associated transitions and can assure safety as compared to a rarely seen transition.

\begin{algorithm}[t]
    \nonl{\textbf{--Data Generation--}}\\
    \nl Assign $D_{train}$ to $\{\}$\;
    \For{$e = 1, 2,\ldots, N$ episodes}
    {
        \For{$t = 1, 2, \ldots, \tau$ timesteps}
        {
            Collect $(s_{t-1},a_{t-1}) \rightarrow s_t$ from the system following policy $\pi$ to generate $D_{train}$\;
        }
    }

    \nonl{\textbf{--Training--}}\\
    \nl Develop a PNN following $(s_{t-1},a_{t-1}) \rightarrow s_t$ for policy $\pi$ over $D_{train}$\;
    \nonl{\textbf{--OOD Monitor--}}\\
    \nl $s_t$ = Current state at timestep $t$ on executing $a_{t-1}$ following policy $\pi$ on system state $s_{t-1}$\;
    $\{s^1_t, s^2_t,\ldots,s^k_t\}$ = set of $k$ predicted current states generated by PNN on feeding $(s_{t-1}, a_{t-1})$\;
    \If{$\left(s_t \in \{s^1_t, s^2_t,\ldots,s^k_t\} \right)$ $\&\&$ $\left( \Pr\left((s_{t-1},a_{t-1}) \rightarrow s_t\right) > \rho \right)$}
    {
        $s_t$ is In-Distribution\;
    }
    \Else
    {
        $s_t$ is Out-Of-Distribution\;
    }    
    \caption{\textit{OOD Monitoring Algorithm($\pi$,$\rho$)}}
    \label{alg:monitor}
\end{algorithm}

\begin{figure*}
    \centering
    \includegraphics[width=\linewidth]{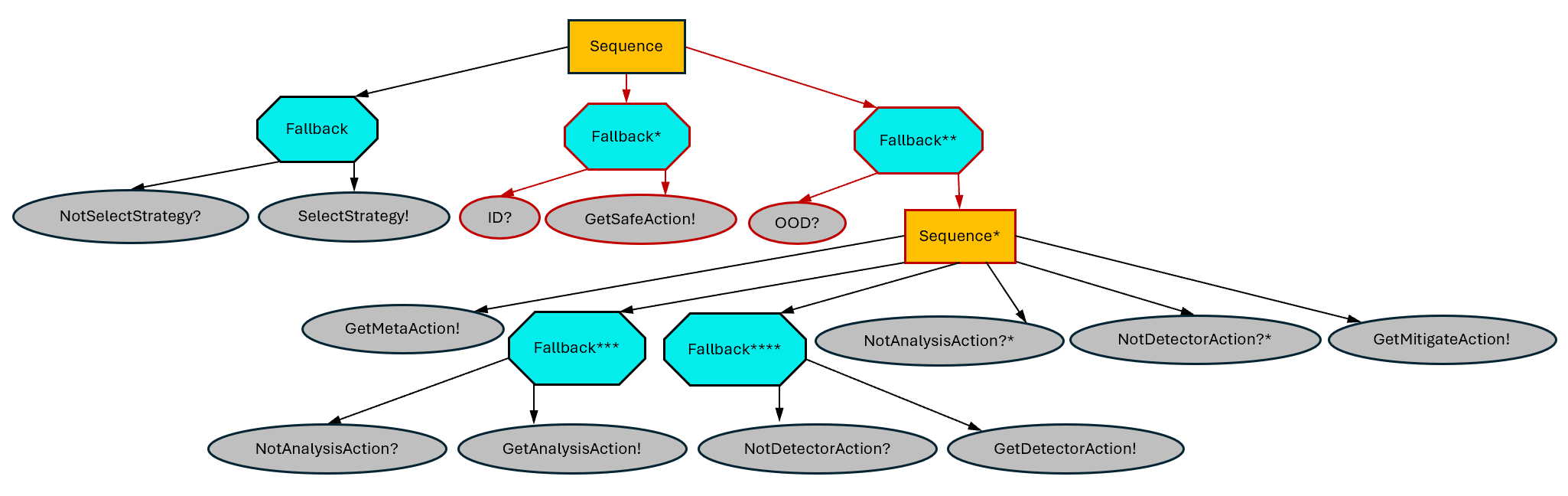}
    \caption{Updated Behavior Tree for OOD Monitoring; The newly added nodes and connections are marked in red in the updated BT.}
    \label{fig:bt_updated}
\end{figure*}

\section{Integration of OOD Monitoring in the EBT}\label{subsec:integration}
\noindent
We need to integrate the safe monitoring behavior of our system in our existing BT so that the system can be monitored at runtime over every tick. To integrate the safe monitoring behavior in the existing BT (shown in Fig.~\ref{fig:gpbt}), we update the hierarchical structure of the BT and add two new $Condition$ behaviors and a new $Action$ behavior in the updated BT. Fig.~\ref{fig:bt_updated} shows our updated BT for OOD monitoring with the newly added nodes and connections marked in red. The new behavior node, $ID?$, stores the PNN for a specific control policy. At each timestep $t$, this node feeds the previous state and action, $(s_{t-1},a_{t-1})$, to the PNN to generate a set of predicted current states, and returns $Failure$ if the current state of the system, $s_t$, is out-of-distribution. The second $Control$ behavior, $OOD?$, returns $Failure$ if the current state of the system, $s_t$, is in-distribution so that the system can continue with the normal operation by executing the remaining part of the BT. We introduce an $Action$ behavior, $GetSafeAction!$, to handle OOD situations. If the current state $s_t$ of our system is OOD, then this node executes $Restore$ action to restore the affected host/server to a previously known ``safe" state, thereby, assuring safety. 

\begin{figure}
    \centering
    \includegraphics[width=\linewidth]{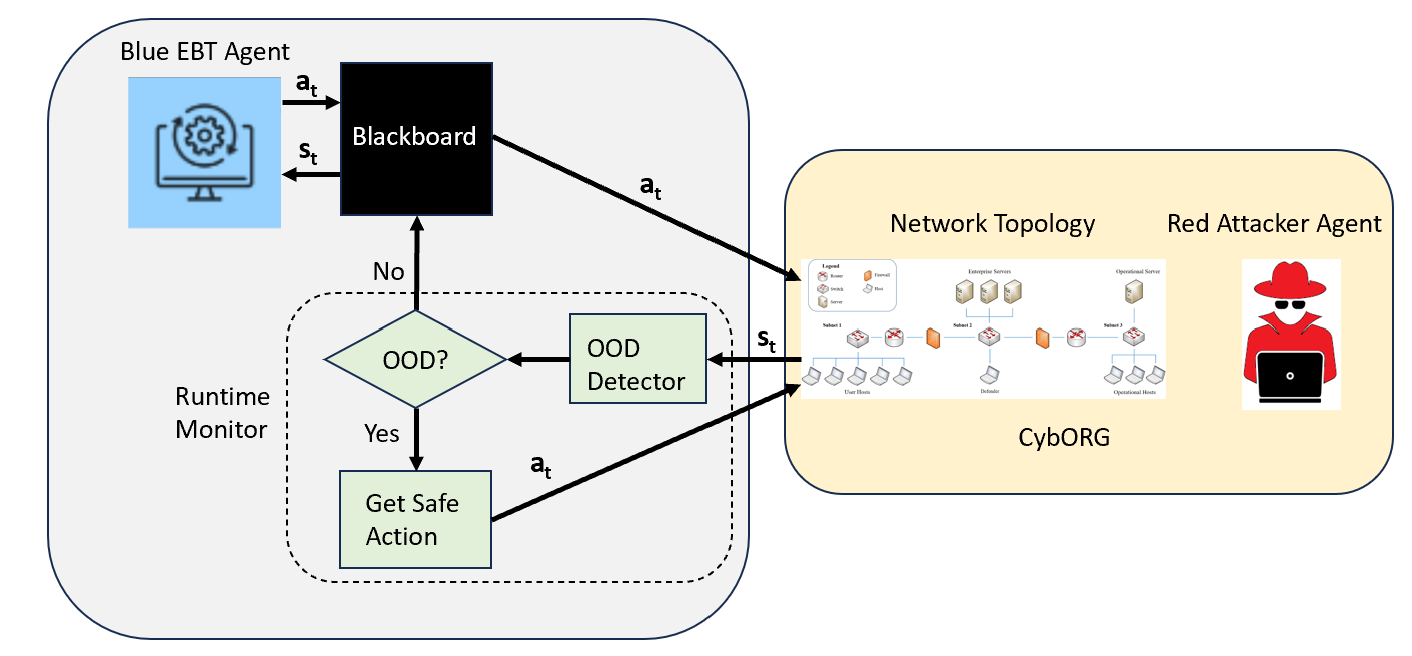}
    \caption{Software Architecture for OOD Monitoring in autonomous cyber-defense environment.}
    \label{fig:arch}
\end{figure}

\section{Experiments and Evaluation}\label{sec:experiment}
\noindent
We execute our proposed OOD Monitoring algorithm for our EBT-based autonomous cyber-defense agent on the simulations of CybORG CAGE Challenge Scenario 2. To demonstrate the effectiveness of our proposed algorithm at runtime, we integrate the OOD monitoring behavior in the EBT  
and evaluate its ability to detect OOD situations online. We conduct our experiments for different Transition Probability Thresholds ($\rho$) under different OOD scenarios with different adversarial strategies as well as adversarial strategy switching.


\subsection{Experimental Setup}\label{subsec:expsetup} 
\noindent
We extend the software architecture of our prior work presented in~\cite{potteiger2024design} to integrate the OOD Monitoring algorithm into our system. Fig.~\ref{fig:arch} shows the extended architecture that executes the EBT-based autonomous cyber-defense agent (Blue EBT agent) and the OOD Monitoring algorithm with the network simulator, CybORG CAGE Challenge Scenario 2. We initialize a blackboard~\cite{pytrees} as the communication interface to communicate between the EBT and the simulator. The blackboard is updated in every iteration to communicate via shared data. To prevent data leaking, the behavior nodes in the EBT cannot directly communicate with the simulator. Instead, they can only access specific data values through this blackboard interface based on their functionality. In each iteration, both the EBT and the simulator can use this blackboard to read and update data values based on their access permissions. We use the PyTrees library~\cite{pytrees} to develop the OOD monitoring behavior in the EBT. Our EBT can detect OOD situations under different attacker strategies as well as when an attacker switches from one strategy to another, and can take safe action whenever such a situation is detected. 

We perform our experiments on a Linux machine with $2.1$~GHz Intel Xeon having $16$ processors and $32$~GB RAM. To generate dataset for training our Probabilistic Neural Network (PNN), we execute our autonomous agent in CybORG CAGE Challenge Scenario 2 under two different adversarial strategies, $B\_line$ and $Meander$, for $10,000$ episodes, each episode consisting of $100$ timesteps. We collect the transitions $(s_{t-1},a_{t-1}) \rightarrow s_t$ in each execution step to generate $D_{train}$. It takes almost $27$ hours to generate the dataset for each agent in CybORG simulator. Note that, each state in CybORG simulator is a vector of $52$ bits where each host/server state is represented with $4$ bits. Two of these bits encode the type of program being executed and the remaining two bits represent the degree to which the host/server has been compromised. Therefore, $2^{4}$ = $16$ distinct bit combinations are possible to represent each host/server state in the network. We observe that only a small subset of these states are reachable. This makes the system scalable as the number of possible reachable states in $D_{train}$ is not very large.
To reduce the computational cost, we label all distinct states in $D_{train}$ and generate the PNN with the labeled states. The full action space of the blue agent is a discrete set of $145$ different actions. We observe that although the dataset generated from the simulator against the $Meander$ red agent has some overlapping states with the dataset generated against the $B\_line$ red agent, the former is much more diverse in nature because of the random behavior of the $Meander$ red agent.
Thus, we generate two PNN, one for each adversarial strategy with the collected datasets. We observe that it takes almost $4$ minutes to train the feedforward PNNs with the collected data during system initialization. To monitor OOD situations at runtime, we update the EBT structure as shown in Fig.~\ref{fig:bt_updated} and discussed in Section~\ref{subsec:integration}.

\subsection{Evaluation under different Transition Probability Threshold}\label{subsec:evaltpt}
\noindent
In our experiments, we use the Transition Probability Threshold ($\rho$) to measure OOD situations. We define an OOD episode as follows.
\begin{definition}
    An episode $e$ is considered to be OOD if there exists at-least one transition in $e$ whose transition probability is less than the Transition Probability Threshold, $\rho$, \emph{i.e.}, $e$ is an OOD episode, if $\exists ((s_{t-1},a_{t-1}) \rightarrow s_t)~|~ \Pr((s_{t-1},a_{t-1}) \rightarrow s_t) < \rho$.
\end{definition}

To evaluate the number of OOD episodes, we execute our OOD Monitoring algorithm with $B\_line$ and $Meander$ as the red adversarial agents and the EBT-based agent as the blue agent over $1000$ episodes, each episode consisting of $100$ timesteps. Table~\ref{tab:oodepisodes} shows the number of OOD episodes for different values of $\rho$. We observe that there are only $1.5\%$ OOD episodes against $Meander$ red agent and $0.1\%$ OOD episodes against $B\_line$ red agent for $\rho$ = $0$. As we increase $\rho$ values gradually from $10^{-5}$, majority of the episodes are OOD. Hence, we set $\rho$ = $0$ for all our subsequent experiments. 

\begin{figure}
    \centering
    \includegraphics[width=\linewidth]{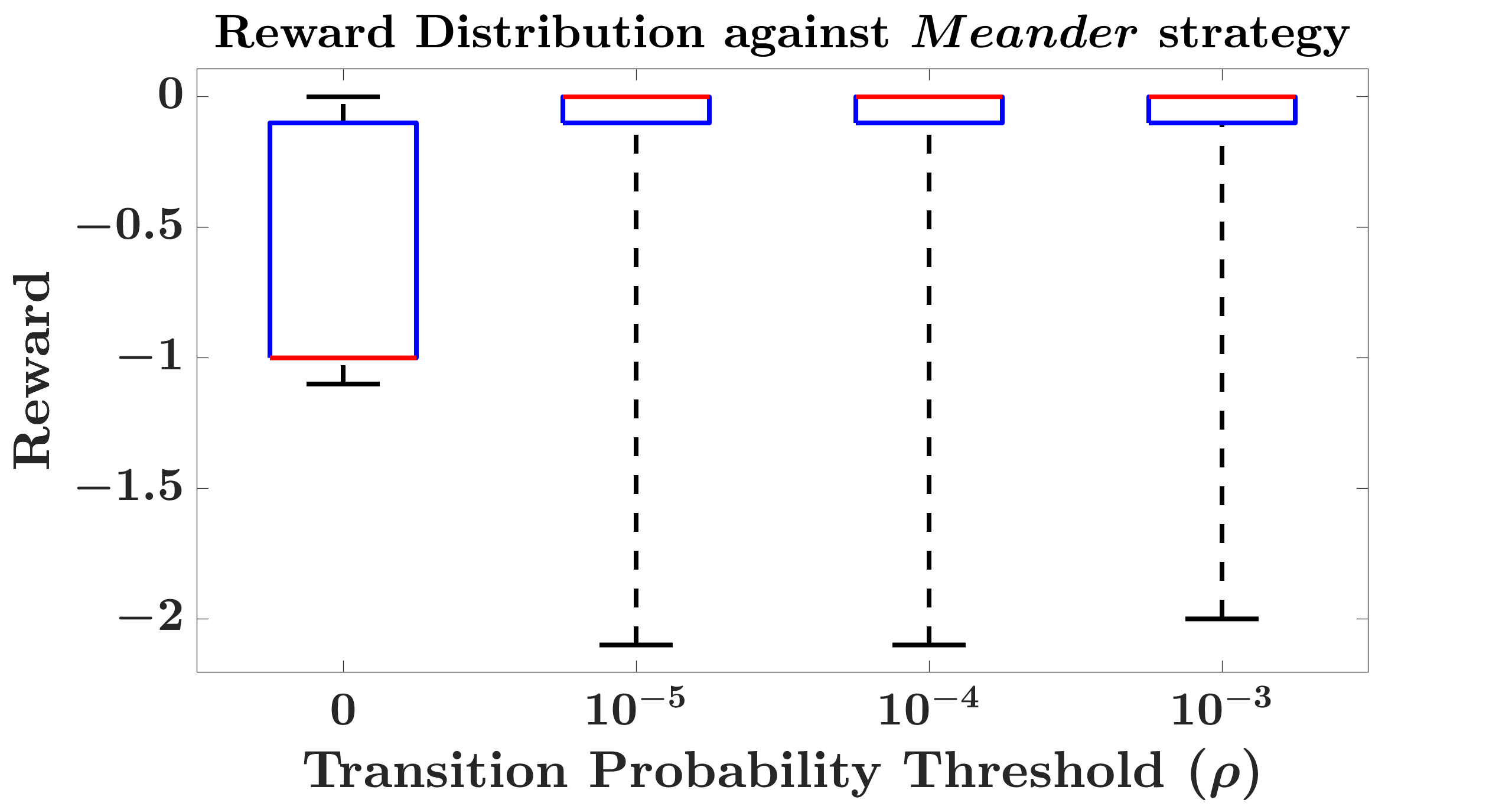}
    \caption{Reward Distribution under different Transition Probability Thresholds ($\rho$) over $1000$ episodes, each with $100$ timesteps against $Meander$ strategy.}
    \label{fig:reward_meander}
    \centering
    \includegraphics[width=\linewidth]{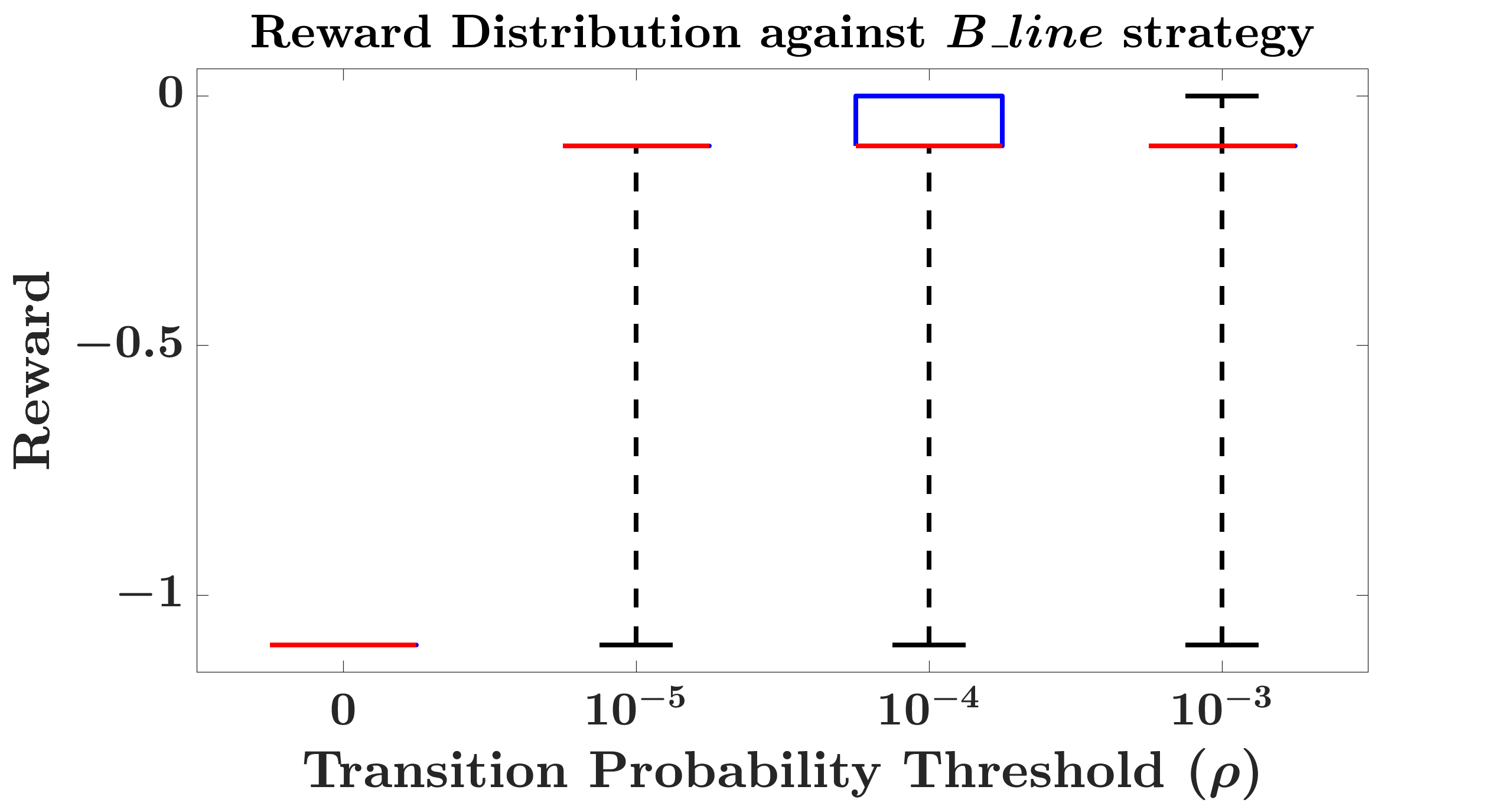}
    \caption{Reward Distribution under different Transition Probability Thresholds ($\rho$) over $1000$ episodes, each with $100$ timesteps against $B\_line$ strategy.}
    \label{fig:reward_bline}
\end{figure}

Fig.~\ref{fig:reward_meander} and Fig.~\ref{fig:reward_bline} show the reward distribution over $1000$ episodes with $100$ timesteps against $Meander$ and $B\_line$ red agents respectively for different values of $\rho$. From Fig.~\ref{fig:reward_bline}, we observe that the reward value is fixed at $-1.2$ against $B\_line$ agent for $\rho$ = $0$. However, the rewards are more diverse against $Meander$ agent for the same value of $\rho$ (refer to Fig.~\ref{fig:reward_meander}). Further, from both the figures, we observe that the median values of the rewards (marked in red in the plots) shift towards $0$ (maximum possible reward for our agents) as we increase $\rho$. This indicates that as we increase $\rho$, we encounter more probable transitions that are known to the system, causing less reward penalties.

\begin{table}[t]
    \caption{Number of OOD episodes with two different Red agent strategies, $Meander$ and $B\_line$, over $1000$ episodes, each with $100$ timesteps}
    \centering
    \begin{tabular}{|c|c|c|}\hline
         Red Agent & Transition Probability & Number of OOD Episodes\\
         Strategy & Threshold ($\rho$) & (out of $1000$) \\ \hline \cline{1-3}
         \multirow{4}{*}{$Meander$}& $0$ & 15 \\ \cline{2-3}
                    & $10^{-5}$& 1000 \\ \cline{2-3}
                    & $10^{-4}$& 1000 \\ \cline{2-3}
                    & $10^{-3}$& 1000 \\ \hline\cline{1-3}
        \multirow{4}{*}{$Bline$}& $0$ & 1 \\ \cline{2-3}
                    & $10^{-5}$& 782 \\ \cline{2-3}
                    & $10^{-4}$& 1000 \\ \cline{2-3}
                    & $10^{-3}$& 1000 \\ \hline
    \end{tabular}
    \label{tab:oodepisodes}
\end{table}

\subsection{Evaluation with EBT}\label{subsec:stratswitch}
\noindent
Apart from uncertainties caused due to limited knowledge about system dynamics at runtime, an OOD situation can also occur due to change in the behavior of an adversarial red agent. To evaluate such situations at runtime, we conduct experiments by integrating the OOD monitoring behavior with the EBT as described in Section~\ref{subsec:runtime}. 

\vspace{0.3em}
\noindent
\emph{Switching to a known adversarial strategy:} We run experiments against $RedSwitch$ strategy where the system instantiates a red agent using $Meander$ strategy and switches to $B\_line$ strategy after a random number of timesteps. We observe that when a red agent switches to a known strategy, \emph{i.e.}, $Meander \rightarrow B\_line$ in our case, our system immediately detects the switch as an OOD situation. However, we observe that we need to restore the state of the system to a previous ``safe" state before switching to a new blue agent policy against the new red agent, \emph{i.e.}, $B\_line$ in our case. Thus, we need the $GetSafeAction!$ behavior in the updated BT (refer to Fig.~\ref{fig:bt_updated}). 

\begin{figure}
    \centering
    \includegraphics[width=\linewidth]{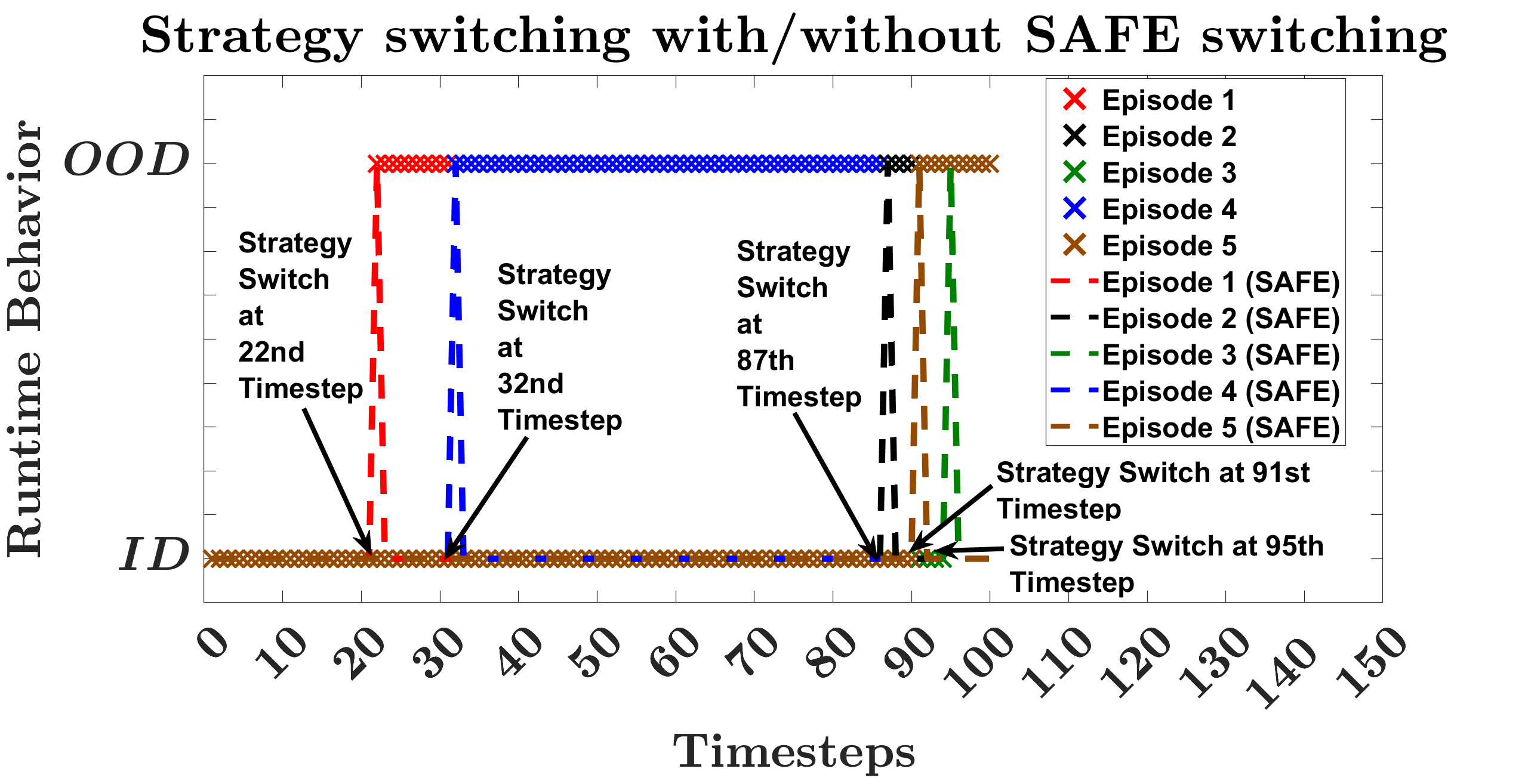}
    \caption{Five episodes under strategy switching ($Meander \rightarrow B\_line$) with $GetSafeAction!$ (plots in dashed lines) and without $GetSafeAction!$ (plots in cross marks) in the EBT.}
    \label{fig:trj_safe}
\end{figure}

To show the necessity of the $GetSafeAction!$ behavior in the updated BT, we conduct two experiments over $1000$ episodes, each with $100$ timesteps. In one experiment we continue with the BT shown in Fig.~\ref{fig:bt_updated}, \emph{i.e.}, with the $GetSafeAction!$ behavior. In the other experiment, we remove this behavior from the BT. Fig.~\ref{fig:trj_safe} shows five random episodes executed in these two setups with the same initial conditions and strategy switching happening at the same timestep in both the cases. The plots marked with dashed lines and cross marks denote the plots under strategy switching ($Meander \rightarrow B\_line$) with and without $GetSafeAction!$ behavior respectively. We observe that when the red agent switches strategy at timestep $t$ (say), the OOD Monitoring algorithm triggers an OOD situation at the immediate next timestep \emph{i.e.}, at timestep $t+1$ in both the cases. In the former setup, \emph{i.e.}, with the $GetSafeAction!$ behavior, the system gets restored to a ``safe" state at timestep $t+1$ and the $SelectStrategy!$ behavior switches the system to the new control policy in the subsequent timestep, \emph{i.e.}, at timestep $t+2$. Thereafter, the system behavior switches back to be in distribution with the new control policy. However, in the latter setup, \emph{i.e.}, without the $GetSafeAction!$ behavior, the system evolves to new unseen states and continues with the OOD situations despite the $SelectStrategy!$ behavior switching the system to the new control policy. 

\begin{figure}
    \centering
    \includegraphics[width=\linewidth]{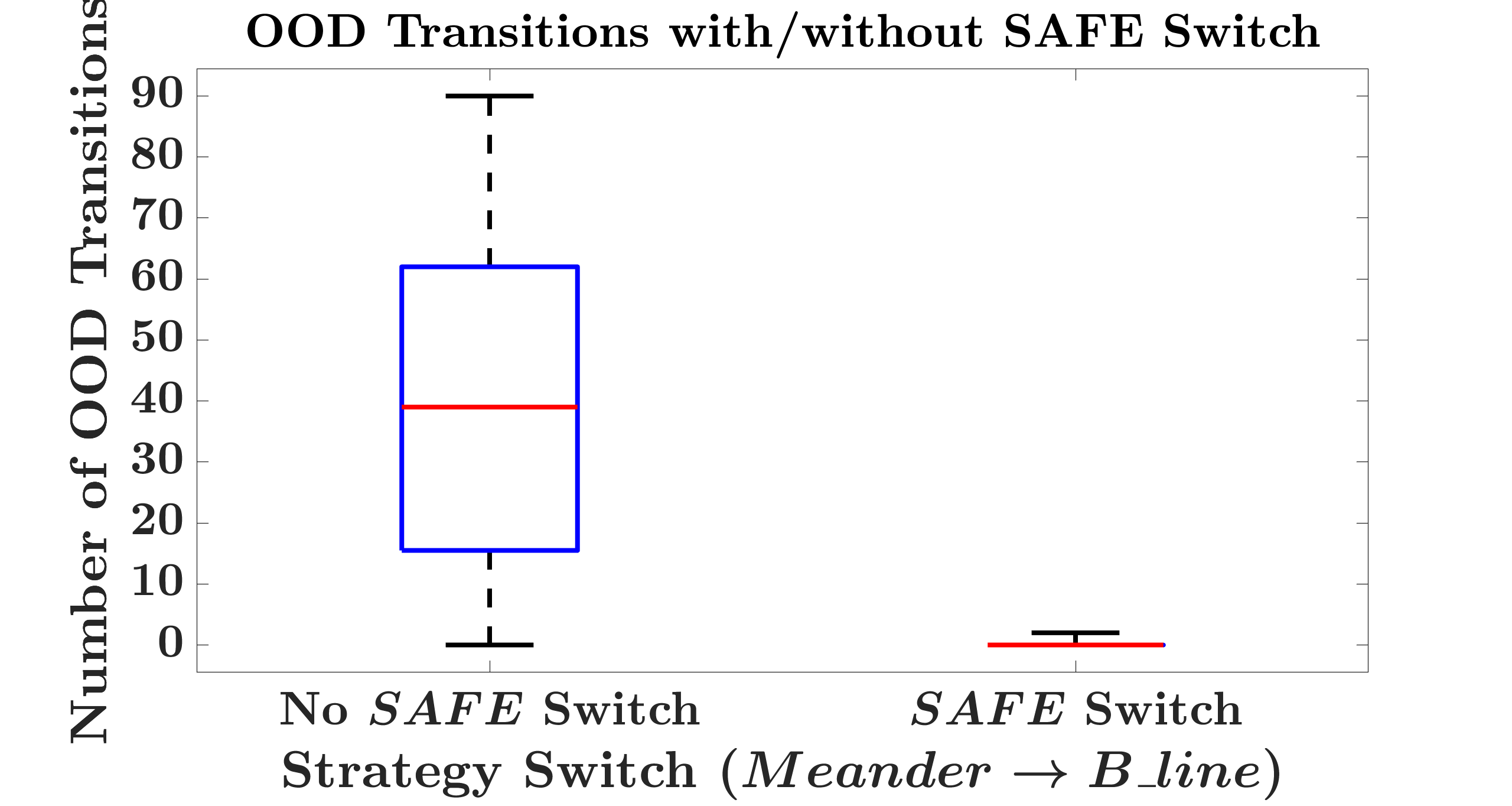}
    \caption{Number of OOD transitions per episode under strategy switching ($Meander$ $\rightarrow$ $B\_line$) without $GetSafeAction!$ behavior (left) and with $GetSafeAction!$ behavior (right) in the EBT.}
    \label{fig:redswitchstat}
\end{figure}

Fig.~\ref{fig:redswitchstat} shows the number of OOD transitions per episode under the two experimental setups, one with the $GetSafeAction!$ behavior and the other without the behavior over $1000$ episodes each with $100$ timesteps. From the figure, we can observe that under ``safe" switching the number of OOD transitions per episode are significantly small and vary between $0$ to $2$, thereby restoring the system back to safe state assuring safety.

\vspace{0.3em}
\noindent
\emph{Switching to an unknown adversarial strategy:} To evaluate the efficiency of our OOD Monitoring algorithm in detecting an unknown red agent strategy, we set only one control policy into the system against $Meander$ and one PNN trained with the transitions generated against $Meander$ red agent. We remove the control policy against $B\_line$ red agent and the corresponding PNN from the system. We also remove the branch of the BT that is responsible for strategy select, \emph{i.e.}, the $NotSelectStrategy?$ and $SelectStrategy!$ behaviors from the EBT. So with this new setup, $B\_line$ red agent strategy serves as an unknown red strategy to our system. Using this setup, we run experiments for upto $1000$ episodes each with $100$ timesteps and compare the results with the one where the system is aware of the $B\_line$ red strategy.

\begin{figure}
    \centering
    \includegraphics[width=\linewidth]{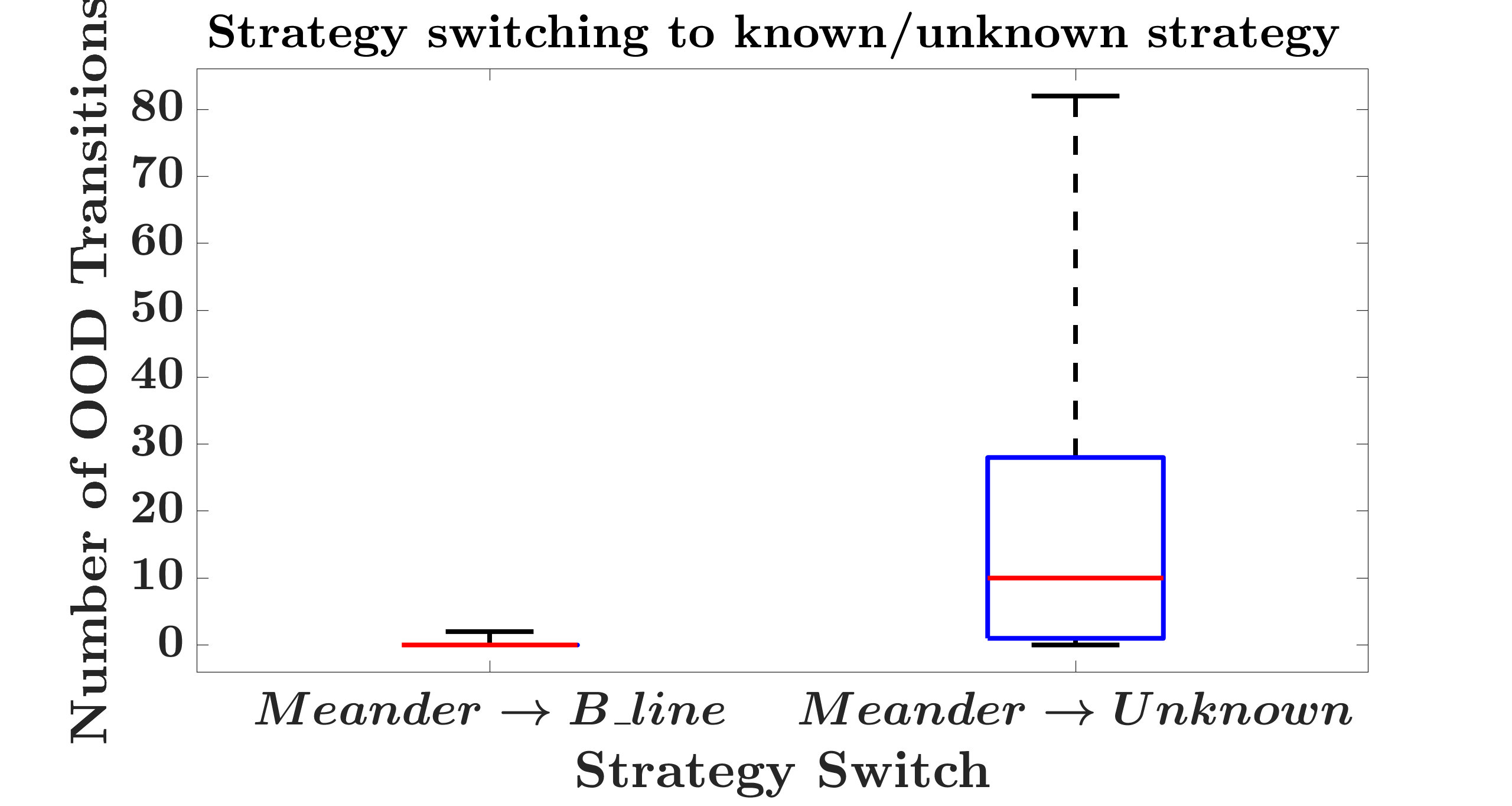}
    \caption{Number of OOD transitions per episode under known strategy switching, $Meander \rightarrow B\_line$ (left), and unknown strategy switching, $Meander \rightarrow Unknown$ (right) in the EBT.}
    \label{fig:stratswitchgeneral}
\end{figure}

Fig.~\ref{fig:stratswitchgeneral} shows the number of OOD transitions per episode where the red agent switches to a known strategy ($Meander \rightarrow B\_line$), and the one where the red agent switches to an unknown strategy ($Meander \rightarrow Unknown$). We observe that under both the conditions, our OOD Monitoring algorithm can promptly detect strategy switching. However, the number of OOD transitions per episode are significantly high when the red agent switches to an unknown strategy as our system is not aware of the states and the transitions associated with the unknown strategy.

\section{Conclusion and Future Works}\label{sec:conclusion}
\noindent
Neurosymbolic cyber-defense agents are increasingly used in autonomous networks to defend complex cyber-attacks. These agents are typically trained with RL policies. However, uncertainties in the runtime environment of these systems pose significant challenges in designing trustworthy agents. These uncertainties arise either due to insufficient knowledge about the runtime dynamics of the system at the time of training these agents, or, due to change in adversarial behavior that remains unknown to the system at training time. To address these challenges, in this work, we propose an OOD Monitoring algorithm that can detect out-of-distribution situations for any RL-based agent with discrete states and actions. To evaluate the performance of our proposed approach at runtime, we integrate it with a neurosymbolic autonomous cyber-defense agent and perform experiments on a complex network simulation environment, the CybORG CAGE Challenge Scenario 2. Experimental results under different adversarial settings show that our proposed algorithm effectively detects OOD situations under all settings and hence can assure safety.

We evaluate our current setting in a simulator. However, the system may act differently and the system dynamics may vary if we execute the same adversarial strategy on a real testbed. In the future, we want to implement our EBT-based autonomous agent on a real emulation environment. Additionally, in our current setting, the adversarial strategy is learnt offline based on the collected dataset from the simulator that characterize the system dynamics at runtime. However, in a realistic scenario, the adversary may switch to a new strategy online. The system needs to adapt and learn the adversarial movements at runtime and enact accordingly. Thus, we want to explore and incorporate online learning techniques to mitigate adversarial attacks on autonomous networks in the future.

\bibliographystyle{ieeetr}
\bibliography{ref}

\begin{thebibliography}{10}

\bibitem{farid2021taskdriven}
A.~Farid, S.~Veer, and A.~Majumdar, ``Task-driven out-of-distribution detection
  with statistical guarantees for robot learning,'' in {\em 5th Annual
  Conference on Robot Learning}, 2021.

\bibitem{shreyas2022}
S.~Ramakrishna, Z.~Rahiminasab, G.~Karsai, A.~Easwaran, and A.~Dubey,
  ``Efficient out-of-distribution detection using latent space of
  $\mathbf{\beta}$-vae for cyber-physical systems,'' {\em ACM Transactions on
  Cyber-Physical Systems}, vol.~6, Apr. 2022.

\bibitem{Cai2020iccps}
F.~Cai and X.~Koutsoukos, ``Real-time out-of-distribution detection in
  learning-enabled cyber-physical systems,'' in {\em 2020 ACM/IEEE 11th
  International Conference on Cyber-Physical Systems (ICCPS)}, pp.~174--183,
  2020.

\bibitem{Colledanchise_2018}
M.~Colledanchise and P.~Ögren, {\em Behavior Trees in Robotics and {AI}}.
\newblock {CRC} Press, jul 2018.

\bibitem{potteiger2024design}
N.~Potteiger, A.~Samaddar, H.~Bergstrom, and X.~Koutsoukos, ``Designing robust
  cyber-defense agents with evolving behavior trees,'' in {\em arxiv}, 2024.

\bibitem{hajdarevic2015pnn}
A.~Hajdarevic, I.~Dzananovic, L.~Banjanovic-Mehmedovic, and F.~Mehmedovic,
  ``Anomaly detection in thermal power plant using probabilistic neural
  network,'' in {\em 2015 38th International Convention on Information and
  Communication Technology, Electronics and Microelectronics (MIPRO)},
  pp.~1118--1123, 2015.

\bibitem{cage_challenge_2}
``Cyber autonomy gym for experimentation challenge 2.''
  \url{https://github.com/cage-challenge/cage-challenge-2}, 2022.
\newblock Created by Maxwell Standen, David Bowman, Son Hoang, Toby Richer,
  Martin Lucas, Richard Van Tassel, Phillip Vu, Mitchell Kiely.

\bibitem{Kiely2023OnAA}
M.~Kiely, D.~Bowman, M.~Standen, and C.~Moir, ``On autonomous agents in a cyber
  defence environment,'' {\em ArXiv}, vol.~abs/2309.07388, 2023.

\bibitem{autonomous2022foley}
M.~Foley, C.~Hicks, K.~Highnam, and V.~Mavroudis, ``Autonomous network defence
  using reinforcement learning,'' in {\em Proceedings of the 2022 ACM on Asia
  Conference on Computer and Communications Security}, ASIA CCS '22, (New York,
  NY, USA), p.~1252–1254, Association for Computing Machinery, 2022.

\bibitem{Wolk2022BeyondCI}
M.~Wolk, A.~Applebaum, C.~Dennler, P.~Dwyer, M.~Moskowitz, H.~Nguyen,
  N.~Nichols, N.~Park, P.~Rachwalski, F.~Rau, and A.~Webster, ``Beyond cage:
  Investigating generalization of learned autonomous network defense
  policies,'' {\em ArXiv}, vol.~abs/2211.15557, 2022.

\bibitem{MolinaMarkham2021NetworkED}
A.~Molina-Markham, C.~Miniter, B.~Powell, and A.~Ridley, ``Network environment
  design for autonomous cyberdefense,'' {\em ArXiv}, vol.~abs/2103.07583, 2021.

\bibitem{neuro2023milcom}
B.~Jalaian and N.~D. Bastian, ``Neurosymbolic ai in cybersecurity: Bridging
  pattern recognition and symbolic reasoning,'' in {\em MILCOM 2023 - 2023 IEEE
  Military Communications Conference (MILCOM)}, pp.~268--273, 2023.

\bibitem{Lundberg1640875}
F.~Lundberg, ``Evaluating behaviour tree integration in the option critic
  framework in starcraft 2 mini-games with training restricted by consumer
  level hardware,'' Master's thesis, KTH, School of Electrical Engineering and
  Computer Science (EECS), 2022.

\bibitem{Li2021MixedDR}
L.~Li, L.~Wang, Y.~Li, and J.~Sheng, ``Mixed deep reinforcement
  learning-behavior tree for intelligent agents design,'' in {\em International
  Conference on Agents and Artificial Intelligence}, 2021.

\bibitem{Bacon2017AAAI}
P.-L. Bacon, J.~Harb, and D.~Precup, ``The option-critic architecture,'' in
  {\em Proceedings of the Thirty-First AAAI Conference on Artificial
  Intelligence}, AAAI'17, p.~1726–1734, AAAI Press, 2017.

\bibitem{Filos2020CanAV}
A.~Filos, P.~Tigas, R.~T. McAllister, N.~Rhinehart, S.~Levine, and Y.~Gal,
  ``Can autonomous vehicles identify, recover from, and adapt to distribution
  shifts?,'' in {\em International Conference on Machine Learning}, 2020.

\bibitem{Averly2023UnifiedOD}
A.~Reza and C.~Wei-Lun, ``Unified out-of-distribution detection: A
  model-specific perspective,'' {\em 2023 IEEE/CVF International Conference on
  Computer Vision (ICCV)}, pp.~1453--1463, 2023.

\bibitem{yangfull2023}
J.~Yang, K.~Zhou, and Z.~Liu, ``Full-spectrum out-of-distribution detection,''
  {\em International Journal of Computer Vision}, vol.~131, p.~2607–2622,
  June 2023.

\bibitem{amaas2023rl}
T.~Haider, K.~Roscher, F.~Schmoeller~da Roza, and S.~G\"{u}nnemann,
  ``Out-of-distribution detection for reinforcement learning agents with
  probabilistic dynamics models,'' in {\em Proceedings of the 2023
  International Conference on Autonomous Agents and Multiagent Systems}, AAMAS
  '23, p.~851–859, International Foundation for Autonomous Agents and
  Multiagent Systems, 2023.

\bibitem{ood2024aamas}
L.~Nasvytis, K.~Sandbrink, J.~Foerster, T.~Franzmeyer, and C.~S. de~Witt,
  ``Rethinking out-of-distribution detection for reinforcement learning:
  Advancing methods for evaluation and detection,'' 2024.

\bibitem{singh2024amaas}
A.~J. Singh and A.~Easwaran, ``Pas: Probably approximate safety verification of
  reinforcement learning policy using scenario optimization,'' in {\em
  Proceedings of the 23rd International Conference on Autonomous Agents and
  Multiagent Systems}, AAMAS '24, p.~1745–1753, International Foundation for
  Autonomous Agents and Multiagent Systems, 2024.

\bibitem{cyborg_acd_2021}
{\em CybORG: A Gym for the Development of Autonomous Cyber Agents}, 2021.

\bibitem{mdp}
M.~L. Puterman, {\em Markov Decision Processes: Discrete Stochastic Dynamic
  Programming}.
\newblock John Wiley \& Sons, 2014.

\bibitem{pytrees}
S.~Reality, ``Pytrees.'' \url{https://github.com/splintered-reality/py_trees},
  2023.

\end{thebibliography}

\end{document}